\documentclass{article}

\usepackage{nac_preprint}
\usepackage[utf8]{inputenc} 
\usepackage[T1]{fontenc}    
\usepackage{hyperref}       
\usepackage{url}            
\usepackage{booktabs}       
\usepackage{amsfonts}       
\usepackage{nicefrac}       
\usepackage{microtype}      

\usepackage{booktabs} 
\usepackage{subcaption}
\usepackage{amsmath}
\usepackage{amssymb}
\usepackage{cleveref}
\usepackage{mathtools}
\usepackage[toc,page]{appendix}
\usepackage{enumitem}
\usepackage{graphics}
\usepackage{graphicx}
\usepackage{xcolor}
\usepackage[export]{adjustbox}
\usepackage{algorithm}
\usepackage{algorithmicx}
\usepackage[noend]{algpseudocode}
\algnewcommand{\LineComment}[1]{\State \(//\) #1}
\algnewcommand{\RLineComment}[1]{\State \(\triangleright\) #1}
\usepackage{bm}
\usepackage{multirow}

\usepackage{wrapfig}

\usepackage{etoolbox}
\usepackage{tikz}
\usetikzlibrary{tikzmark}
\usetikzlibrary{calc}

\errorcontextlines\maxdimen

\newcommand{\ALGtikzmarkcolor}{black}
\newcommand{\ALGtikzmarkextraindent}{4pt}
\newcommand{\ALGtikzmarkverticaloffsetstart}{-.5ex}
\newcommand{\ALGtikzmarkverticaloffsetend}{-.5ex}
\makeatletter
\newcounter{ALG@tikzmark@tempcnta}

\newcommand\ALG@tikzmark@start{%
    \global\let\ALG@tikzmark@last\ALG@tikzmark@starttext%
    \expandafter\edef\csname ALG@tikzmark@\theALG@nested\endcsname{\theALG@tikzmark@tempcnta}%
    \tikzmark{ALG@tikzmark@start@\csname ALG@tikzmark@\theALG@nested\endcsname}%
    \addtocounter{ALG@tikzmark@tempcnta}{1}%
}

\def\ALG@tikzmark@starttext{start}
\newcommand\ALG@tikzmark@end{%
    \ifx\ALG@tikzmark@last\ALG@tikzmark@starttext
    \else
        \tikzmark{ALG@tikzmark@end@\csname ALG@tikzmark@\theALG@nested\endcsname}%
        \tikz[overlay,remember picture] \draw[\ALGtikzmarkcolor] let \p{S}=($(pic cs:ALG@tikzmark@start@\csname ALG@tikzmark@\theALG@nested\endcsname)+(\ALGtikzmarkextraindent,\ALGtikzmarkverticaloffsetstart)$), \p{E}=($(pic cs:ALG@tikzmark@end@\csname ALG@tikzmark@\theALG@nested\endcsname)+(\ALGtikzmarkextraindent,\ALGtikzmarkverticaloffsetend)$) in (\x{S},\y{S})--(\x{S},\y{E});%
    \fi
    \gdef\ALG@tikzmark@last{end}%
}

\apptocmd{\ALG@beginblock}{\ALG@tikzmark@start}{}{\errmessage{failed to patch}}
\pretocmd{\ALG@endblock}{\ALG@tikzmark@end}{}{\errmessage{failed to patch}}
\makeatother

\usepackage[most]{tcolorbox}
\definecolor{timecolor}{RGB}{0,90,160}
\definecolor{resultcolor}{RGB}{0,120,70}
\definecolor{softgreen}{RGB}{245,252,247}
\newtcolorbox{resultbox}{
  colback=softgreen, colframe=resultcolor!75!black,
  boxrule=0.8pt, arc=2mm,
  left=2mm, right=2mm, top=1mm, bottom=1mm}

\newcommand{\Normal}{\mathcal{N}}

\DeclarePairedDelimiter{\norm}{\lVert}{\rVert}
\DeclarePairedDelimiter{\abs}{\lvert}{\rvert}
\newcommand{\layerWidth}{N}      
\newcommand{\vct}[1]{\bm{#1}}                           
\newcommand{\vect}[1]{{\vct{#1}}}                       
\newcommand{\colvect}[1]{{\underline{\vct{#1}}}}        
 
\newcommand{\latent}[1]{\vect{z}_{#1}}    
\newcommand{\obs}{\vect{o}}      
\newcommand{\target} {\vect{y}}
\newcommand{\latents}{\colvect{z}}       
\newcommand{\error}[1]{\vect{\epsilon}_{#1}}
\newcommand{\synaps}{W}                   
\newcommand{\actf}{\phi}                
\newcommand{\lract}{\lambda}
\newcommand{\nlayers}{L}
             
\newcommand{\pred}[1]{\vect{\mu}_{#1}}

\title{Error Highways: Scaling Predictive Coding to Very Deep Networks}

\author{%
Amirhossein Mohammadi \\
SingularityNET\\
Zug, Switzerland \\
\texttt{amir.mohammadi@singularitynet.io}
\And
Alexander G. Ororbia \\
Rochester Institute of Technology\\
Rochester, NY, USA \\
\texttt{ago@cs.rit.edu}
}

\begin{document}

\setlength{\abovedisplayskip}{0.065cm}
\setlength{\belowdisplayskip}{0pt}

\maketitle

\begin{abstract}
Predictive coding networks (PCNs) offer a biologically-plausible, local-learning alternative to back-propagation of errors (backprop). Nevertheless, they have remained largely confined to shallow architectures and evaluated on simple machine intelligence benchmarks. A central obstacle to scaling PCNs is that the learning signal decays rapidly as it propagates away from the clamped boundaries, leaving interior layers effectively unchanged.
To directly counter this problem, we propose \textbf{highway error propagation} (HEP), a scheme that augments the free energy function underlying predictive coding (PC) by altering its neural structure with
feedback matrices $V_{\nlayers\to i}$ that couple selected hidden states directly to the clamped output error.
Since this coupling is linear in the hidden state, the highway pathway delivers a correction at every inference step whose magnitude is \emph{independent of depth}, in contrast to vanilla PC where the output error reaches the $i$-th hidden layer with attenuation that decays exponentially in depth. This bypasses the Jacobian chain while preserving the local PC synaptic update rule. On MNIST and Fashion-MNIST, we show that HEP effectively trains MLPs of up to $128$ layers with accuracy that is robust with respect to depth.

\end{abstract}

\section{Introduction}
\label{sec:intro}

Backpropagation of errors (BP) is the workhorse behind modern-day deep learning; it is the standard way of training artificial neural networks (ANNs). However, even though it has proven to be effective for conducting credit assignment \cite{rumelhart1986learning}, it depends on a global feedback pathway to transmit error signals backwards along through an entire network of neurons (i.e., back along the forward propagation pathway taken during inference).
This backward pass is not only biologically implausible but it is non-local given that computing each layer's update requires information obtained from the rest of the network, not just adjacent layers~\cite{ororbia2023brain}.

As a result of the above, there is growing interest in biologically-inspired learning rules and credit assignment schemes that are local; these specifically seek to adapt the parameter values associated with each neuron or layer  using only the signals available to those particular neurons or layers.
Predictive coding (PC) is one of the most prominent of these learning schemes~\cite{rao_ballard,friston2005,salvatori2025survey}. In a PC circuit, each layer of neurons predicts the activity of those of the layer beneath (where the ``bottom'' of the circuit is taken to be the sensory input layer), and learning proceeds by reducing the resulting prediction errors with purely local updates.
This gives a clean, local solution to the credit assignment problem~\cite{alex_ngc,ororbia2023brain}. Aside from biological plausibility, PC also has been demonstrated to offer better performance over BP-trained ANNs in online learning and continual learning settings \cite{ororbia2020continual,ororbia2022lifelong,salvatori2024stable,salvatori2025survey}, showing that PC offers a serious alternative to BP due to its performance advantages.

Moreover, the scaling of BP-trained models is one of the central reasons behind their success in the current machine learning (ML) era. Deep neural networks (DNNs) tend to learn richer features as well as complex mappings. If PC is to be a serious alternative to BP-based DNNs, it should also  scale at least as well as a BP-trained DNN does and, furthermore, solve similar, complex tasks. However, in practice, PC has still failed to scale methodically; primary PC-related work has been restrained to small, fully-connected layers or a small handful of convolutional layers while deeper architectures, e.g., residual network structures \cite{resnet}, remain out-of-reach. This is the gap that prevents PC from being useful on the complex, larger-scale problems that BP-based systems succeed on.

The main obstacle for scaling a PC network (PCN) -- a form of PC \cite{whittington_bogacz} designed to map a sensory input to a target output (such as a label or target value) -- to a deep architecture is that the learning signal in PC weakens as it travels through the network. In effect, PC is a message-passing scheme where, at each inference step, error information (i.e., the learning signal) is passed from one layer to the next. The issue arises when the inference procedure begins, where the learning signal tries to reach a part of the PCN that is $k$ ``hops'' away from the source of learning signal (such as the output layer) in order to update neural activity states. However, hop by hop, the signal attenuates by the (neural) activity step size $\lract$ to the point that, when it travels $k$ hops along the network, it has already been attenuated by $\mathcal{O}(\lract^k)$ \cite{ePC}. This results in layers that are away from the learning signal sources receiving little to no updating signal; this means that they stay relatively close to their initial state value(s). Note that this ``vanishing'' residual/error signal is similar in spirit to the problem of vanishing gradients in BP-based ANNs \cite{bengio2014auto,ororbia2023brain}.

To overcome the above attenuation problem, we introduce \textbf{highway error propagation} (HEP), a credit assignment scheme that takes advantage of PC's inherent message-passing neuronal structure and entangled form of inference-and-learning \cite{alex_ngc,salvatori2025survey}; note that our scheme's naming pays homage to the use of the long-range, forward skip-connections once employed to train classical recurrent neural networks \cite{srivastava2015highway,zilly2017recurrent}, in the era before generative pre-trained transformer models. HEP specifically adds a set of message-passing, feedback matrices $V_{\nlayers\to i}$ that carry a supervisory signal straight to a set of selected hidden layers within a single hop. This coupling is not only more neurobiologically realistic (as  networks of biological neurons are heavily recurrently connected, often in a more heterarchical fashion \cite{singer2021recurrent,ororbia2023brain}), but it also crucially delivers a correction at every inference step whose size does not depend on the depth.
In effect, the early layers are pushed away from their initial values from the very first step of inference/message-passing instead of waiting for a signal that has already decayed along on the way. Notably, HEP only augments the usual PC inference loop and leaves the standard local PC synaptic weight update unchanged.

\noindent
\textbf{Contributions.} In this work, we make the following key contributions:
\begin{itemize}[noitemsep,nolistsep]
\item {we provide a unified account of the depth problem in PCNs as two failures across different time scales, i.e., in terms of a transient and a steady-state decay of the learning signal;}
\item  {guided by this view, we propose highway error propagation (\textbf{HEP}), a scheme that delivers the output error directly to hidden layers in a single hop, enabling the training of very deep PCNs; and}
\item \textcolor{black}{we isolate HEP's effect with controlled experiments: PCNs have
no forward skip connections, the PCN baseline is identical with $\alpha=0$, and a sweep of $\alpha$ traces a transition from chance to stable
training (up to $128$ layers).}
\end{itemize}
The rest of the paper is organized as follows. Section~\ref{sec:background} reviews PC and sets up our notation. Section~\ref{sec:depthProb} explains the depth problem in PCNs as steady-state and transient decay of the learning signal. Section~\ref{sec:related} covers existing remedies for deep PC models and the gap that HEP fills. Section~\ref{sec:method} presents HEP and its update rules, and Section~\ref{sec:experiments} reports our experiments.
\section{Background: Predictive Coding}
\label{sec:background}

In this work, we consider a PCN -- which is, as mentioned earlier, a supervised format of predictive coding -- composed of $\nlayers$ layers; all layers are represented as activity states
$\latents=(\latent{0},\latent{1},\dots,\latent{\nlayers})$. For each layer, a feedforward prediction $\pred{i}=f_{i-1}(\latent{i-1})$ is made of a nearby layer/state (technically the one below it).
The two extreme ends of the PCN (the input layer $i=0$ and the output layer $i=\nlayers$) are clamped to the data and target values; in other words, the input layer is clamped to an observation, i.e., $\latent{0}\equiv\obs$, at the bottom while the top layer is clamped to the target, i.e., $\latent{\nlayers}\equiv\target$. Holding these two layers fixed at their clamped values, inference minimizes the energy over the free interior states in accordance with the energy functional below:
\begin{equation}
  \label{eq:energy}
  E(\latents) \;=\;
  \underbrace{\sum_{i=1}^{\nlayers-1}\tfrac{1}{2}\norm{\latent{i}-\pred{i}}^2}_{\text{prediction errors}}
  \;+\;
  \underbrace{\ell\!\left(\latent{\nlayers},\pred{\nlayers}\right)}_{\text{supervisory signal}},
\end{equation}
where $\ell$ is any supervisory loss that measures the mismatch between the output
prediction $\pred{\nlayers}$ and the clamped target value $\latent{\nlayers}$. The hidden prediction errors and the supervisory output error are formulated as follows:
\begin{equation}
  \label{eq:errors}
  \error{i} := \latent{i}-\pred{i}\quad(i=1,\dots,\nlayers-1),
  \qquad
  \error{\nlayers} := -\,\nabla_{\pred{\nlayers}}\,\ell\!\left(\latent{\nlayers},\pred{\nlayers}\right).
\end{equation}
Note that the output error $\error{\nlayers}$ is the only signal that carries target (e.g., label) information into the PCN network.

For a PCN, we initialize the initial conditions of the inference (i.e., the initial values of each non-input/output layer) with a single feedforward sweep, setting every free state to the prediction that it receives from the layer below; this ``warm start'', or ancestral projection \cite{ororbia2023brain}, places the network near a low-energy configuration, providing a good, message-passing-friendly initialization~\cite{FasterPCN}.
Gradient descent on the (free) energy $E$ then refines these states, with activity step size $\lract$ and layer Jacobian $J_i=\partial f_i/\partial\latent{i}$, carried out in the following manner:
\begin{equation}
  \label{eq:z_update_pc}
  \latent{i}^{(t+1)}
  = \latent{i}^{(t)}-\lract\,\nabla_{\latent{i}}E
  = \latent{i}^{(t)}-\lract\bigl(\error{i}^{(t)}-J_i^{(t)\top}\error{i+1}^{(t)}\bigr).
\end{equation}
The above is effectively the design an expectation step (E-step) in the expectation-maximization, inference-and-learning scheme  \cite{dempster1977maximum,salvatori2025survey} that characterizes PCNs and PC more generally.
Note that, while the E-step is generally carried in a fashion similar to Equation \ref{eq:z_update_pc}, or in the framework of differential equation (Euler) integration \cite{alex_ngc}, one may employ more advanced updating schemes, such as Adam \cite{adam}, as we do in this work.
Once the neural states settle (usually to a steady-state value, found after carrying out many E-steps), we then update the synapses, i.e., the maximization step (M-step),  using the local (Hebbian) update rule:
\begin{equation}
  \label{eq:W_update_pc}
  \Delta W_\ell \;\propto\;
  \bigl(\error{\ell+1}\odot\actf'(W_\ell\latent{\ell}+b_\ell)\bigr)\latent{\ell}^{\top}.
\end{equation}
Both of the above updates are local (in both space and time); crucially, a layer uses only its own error and the error of its
neighbor. This locality, a key characteristic that lends PC credit assignment its biological plausibility, is the property that we seek to preserve in Section~\ref{sec:method}.

\section{The Problem of Depth in Predictive Coding}
\label{sec:depthProb}

Scaling PCNs to many layers has been the focus of several recent efforts~\cite{muPC,metaPCN,ePC,precisionGuided}. The obstacles that these studies identify all concern PCN inference dynamics, particularly in terms of where the free states settle once the free energy has been minimized as well as how the neural states behave along the way. We categorize these two points of focus into steady-state and  transient (regime) forms of analysis, and refer the reader to Appendix~\ref{app:depth} for the full derivations.

\subsection{PCN Dynamics: Steady-State Analysis}
\label{sec:steady}

At equilibrium, a PCN's free states stop ``moving'', such that $\nabla_{\latents}E=0$; notably, the layer-wise condition unrolls into a product of Jacobians as follows:
\begin{equation}
  \label{eq:steady}
  \error{\ell}=J_\ell^{\top}\error{\ell+1}
  \quad\Longrightarrow\quad
  \error{i}=\Bigl(\prod_{k=\nlayers-1}^{\,i}J_k\Bigr)^{\!\top}\error{\nlayers} .
\end{equation}
This recovers the familiar result that a PCN reproduces  backprop's error signals at equilibrium~\cite{whittington_bogacz,millidge2022backprop}. If we take the spectral norms of~\eqref{eq:steady}, the error that reaches layer $i$ is bounded by the product of the Jacobian norms above it, i.e., $\norm{\error{i}}\le\bigl(\prod_{k=i}^{\nlayers-1}\norm{J_k}\bigr)\norm{\error{\nlayers}}$. When these norms stay below $1$, then this signal decays geometrically with depth; when these norms stay above $1$, the signal explodes~\cite{metaPCN}.

\subsection{PCN Dynamics: Transient Analysis} %
\label{sec:transient}

Note that the picture above assumes that inference has reached equilibrium. In particularly deep networks, this assumption is fragile. The inference problem becomes increasingly ill-conditioned with depth, such that each step is smaller and convergence slower as the network becomes deeper \cite{muPC}.
Under a fixed inference (computational) budget, the states are often stopped well before equilibrium; as a result, it matters how these states move early on in the trajectory induced by the inference dynamics. Under feedforward initialization, only the output error is active at time-step $t=0$. This error travels back one layer per step and picks up a factor of $\lract$ at each hop; thus, the first update to reach layer $i$ is~\cite{ePC}:
\begin{equation}
  \label{eq:transient}
  \Delta\latent{i}\;\propto\;\lract^{\,\nlayers-i}\,\error{\nlayers}^{(0)} .
\end{equation}
Since $\lract$ is small for stability, early layers receive almost no supervisory signal to drive their updates and tend to 
sit at their initial feedforward values.

\begin{figure}[t]
  \centering
  \includegraphics[width=1\linewidth]{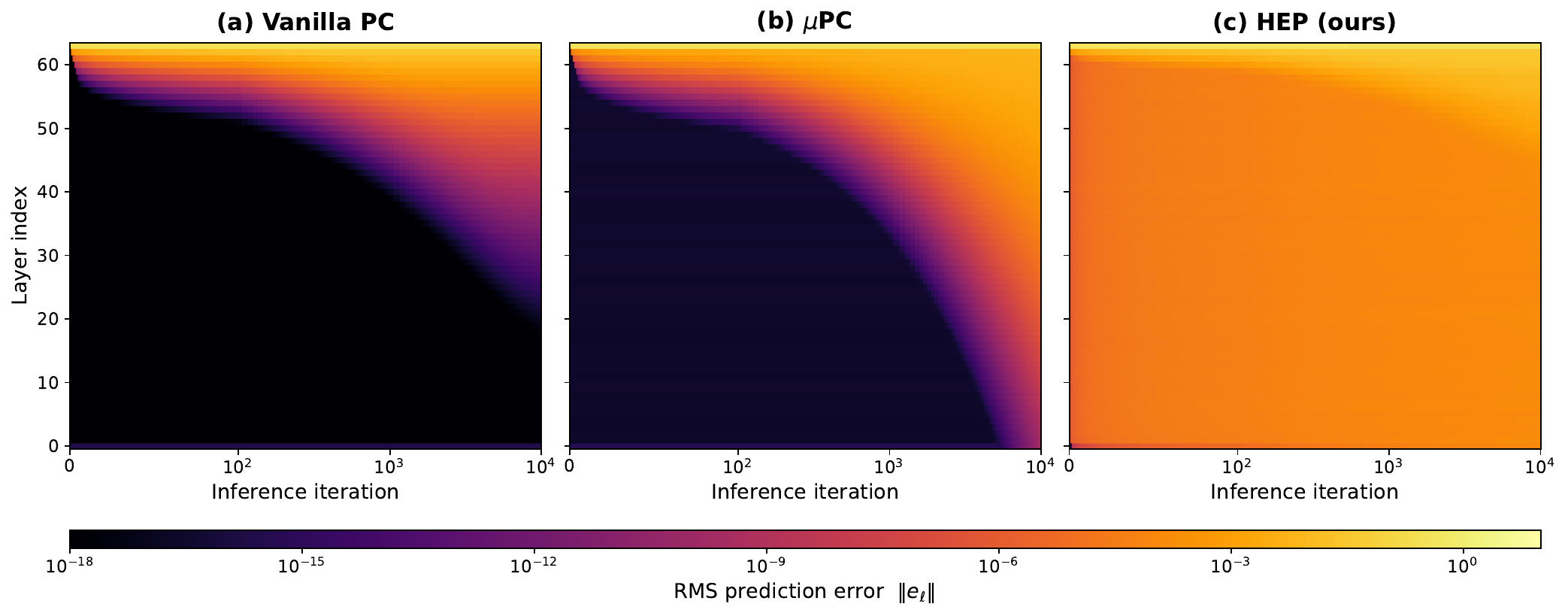}
  \caption{
  \textbf{Per-layer RMS prediction error $\Vert e_\ell \Vert$ (color) versus layer ($0=$ input) and inference iteration for a depth-64 MLP.}
  In \textbf{(a)} vanilla PC and \textbf{(b)} $\mu$PC the output
  error only crawls back along the network, stalling deep and leaving the
  input-side layers at the numerical floor; \textbf{(c)} HEP's error highways
  inject it directly into the hidden layers, guiding the learning procedure early in inference.}
  \label{fig:energy-propagation}
\end{figure}

\section{Extant Remedies to the Problem of Depth}
\label{sec:related}
In this section, we revisit key prior work that is related to our effort in attempting to tackle the
problem of scaling PCNs. In order to solve the challenges of Section~\ref{sec:depthProb}, one must either:
(1) change the optimization that solves for the (PCN) states or,
(2) change the PCN architecture and its wiring (i.e., its structure).

Notably, in the spirit of the first potential way to solve the depth problem, ePC~\cite{ePC} and Meta-PCN~\cite{metaPCN} both change the optimization mechanism. ePC specifically opts for a change of variables in the inference (without moving the minima of the free energy) and employs a preconditioner on the Hessian of the optimization in order to make the problem better conditioned. However, the main issue with ePC is that it relies on non-local information which violates our desired criterion of update locality to ensure biological plausibility. In contrast, Meta-PCN reformulates the  optimization problem entirely; instead of minimizing the energy, it minimizes a linearization of the norm of its
gradient, i.e., $\norm{\nabla_{\latents}E}^2$.

Alternatively, $\mu$PC~\cite{muPC} rescales the weights with a depth-aware parameterization such that the PCN neural activities stay well-behaved as network depth grows. This stabilizes the forward pass but it does \emph{not} address the vanishing learning signal and the
error that reaches the early layers will still be tiny/negligible. In Figure~\ref{fig:energy-propagation}(b), we  show this directly; here, under $\mu$PC, we see that the error requires thousands of inference (E-step) iterations to even arrive at the early layers. This means that, within a practical, often resource-constrained budget, the layers in between receive almost no learning signal. Despite this issue, $\mu$PC nevertheless trains MLPs of over $100$ layers on MNIST.

We remark that the above works because $\mu$PC introduces (feed)forward skip connections, which were originally and historically employed in some early generative PC models \cite{alex_ngc}, that carry a copy of the input (or state) straight to the final layers, ensuring that the network can reach good accuracy while the layers in between are hardly updated; in a sense, this is reminiscent of deep residual networks \cite{resnet} or DNNs with short-cut pathways (however, the layers in the case of $\mu$PC would just not be utilized, in general).
This is a reasonable outcome for the study's strong performance at depth \cite{muPC} but it leaves a different question open; namely, \emph{is it still possible to construct a scheme where every layer of a deep PCN is utilized and receives a learning signal and learn?}
This motivates this current work and our proposed solution and will be the question that we target. As a result, our architecture and scheme will not utilize forward skip connections at all (though it is entirely possible for one to combine this mechanism with our PCN models if desired). Our experimental goal focuses on the constraint that the only path(way) from input to output in a PCN runs through
every layer; this means that the network will struggle to perform well unless the interior layers learn to extract meaningful/reasonable features.
Note that another line of work~\cite{precisionGuided} accelerates the signal traveling back through the PCN by multiplying the error with a larger precision matrix, such that the transient decay of Section~\ref{sec:transient} is reduced. This work also stabilizes the forward pass with a frozen form of batch normalization which normalizes the neural activities during inference.

More broadly, in the realm of biologically-inspired credit assignment and neuroscience-informed artificial intelligence (NeuroAI) \cite{salvatori2025survey,zador2026neuroai}, local representation alignment (LRA)~\cite{ororbia2019biologically,rec_lra}, direct feedback alignment (DFA)~\cite{nokland_dfa}, and direct Kolen-Pollack feedback alignment (DKP)~\cite{DKP_PC} are the
closest efforts to ours in terms of pursuing the second pathway to solving the problem of depth, i.e., altering the wiring and structure. These approaches, even though they are often applied to different, non-PCN neural architectures, all integrate synaptic connections from the output layer
directly to the hidden layers.
DFA~\cite{nokland_dfa} uses fixed random feedback matrices and then updates the weight matrices in one pass, with no backward chain. Recursive LRA (Rec-LRA) similarly uses
local feedback to produce a target for each layer and updates the weights to match
it in one shot; it also investigates the recursive application of local, internal short-circuit pathways to ensure the training of blocks of neural operators (e.g., the set of convolutions and pooling mechanisms of a residual block). Nevertheless, neither Rec-LRA or DFA are based in PC and neither runs an inference loop.
DKP-PC is the most related, and is the only one built on PC. Before inference, DKP-PC runs one preliminary weight update at every layer, using a learned feedback
projection of the output error, which makes the prediction errors non-zero everywhere from the start (of inference) such that a single inference step is ample.
However, the feedback enters only through that one weight step and the inference update itself remains identical to the ordinary PC format. Thus, the signal that actually reaches the middle PCN layers is a single ``kick'' and any further guidance from the output error must travel back through the usual chain, in effect decaying with depth much as in vanilla PC.

The proposed framework in our study, highway error propagation (HEP), instead injects the current output error into the hidden states at every inference step, such that the correction stays depth-independent throughout inference and keeps acting until the (PCN) states settle. Moreover, HEP's crucial mechanistic difference with DKP-PC is notably inspired by the brain's heavy recurrent connectivity \cite{pessoa2023entangled,hawkins2025hierarchy}, and the fact that PC is inherently an intertwined inference-and-learning model (which means that the learning/plasticity of the synapses are affected by the structure/architecture of the model's inference \cite{ororbia2023brain,salvatori2025survey}), of which has been postulated to play an important role in learning useful distributed representations of sensory input in computational neuroscience PC models \cite{ororbia2020continual,alex_ngc} (these last efforts have argued for heterarchical formulations of PC).
Furthermore, DKP-PC has only been empirically demonstrated to work with shallow models, a $3$-layer MLP and a $9$-layer VGG; these are far shallower than the depths where the attenuation described in Section~\ref{sec:transient} becomes severe.
In contrast, our experiments will target depths of $32$ to $128$ layers, where the problem of depth and decaying error signals cannot be avoided.

\section{Methodology}
\label{sec:method}

Following the transient analytical picture that we presented in Section~\ref{sec:transient}, the interior of a PCN is free of error right after the feedforward initialization and
the output error reaches layer $i$ only after $\nlayers-i$ inference (E-)steps. By the time the error gets to layer $i$, it has been further attenuated to $\mathcal{O}(\lract^{\,\nlayers-i})$. So within a finite, resource-constrained inference budget, the supervisory signal only reaches a handful of layers close to the output, while
the remaining hidden layers stay at their feedforward values throughout the duration of the full inference process.

This motivates our proposed HEP scheme. Instead of letting the output error take this long to traverse an inefficient pathway to the hidden layers, we deliver the supervisory signal to a subset of hidden layers at the very beginning of inference; this early, neurally-plausible delivery will guide
the activities of the internal layers/states from the start. Specifically, we do this by generalizing the message-passing mechanics of a PCN by introducing a synaptic feedback matrix that carries the error to a particular hidden layer in a single hop. Under HEP, the correction or ``nudge'' (\cite{millidge2022backprop}) that each hidden layer receives no longer depends on the depth of the network; this means that these layers will be updated regardless of how deep they sit within the PCN structure.

Again, we point out that HEP has a similar spirit to a residual connections \cite{resnet} or highway short-circuit lines \cite{srivastava2015highway,zilly2017recurrent}, but it acts on a different
pathway -- specifically, it modifies feedback transmission that underwrites the message-passing of errors.
A ResNet~\cite{resnet} specifically adds shortcuts in the (feed)forward pass so that, during training, the learning/teaching signal does not have to pass through every layer
in order to reach an early one.
In a similar vein, we add a shortcut message-passing pathway from the last (output) layer such that the supervisory error does not have to hop through many nodes to reach a
hidden layer (such as one close to the sensory input). It is important to point out that forward skip connections can be added on top of our architecture, but the essential part that enables learning in our PCN is HEP (hence our focus on constraining the forward pathway of the PCN to be skip-connection-free in this work) since it is what will carry the learning/mismatch signal back to the early hidden layers.

\begin{figure}[!t] 
  \centering
    \centering
    \includegraphics[width=1\linewidth]{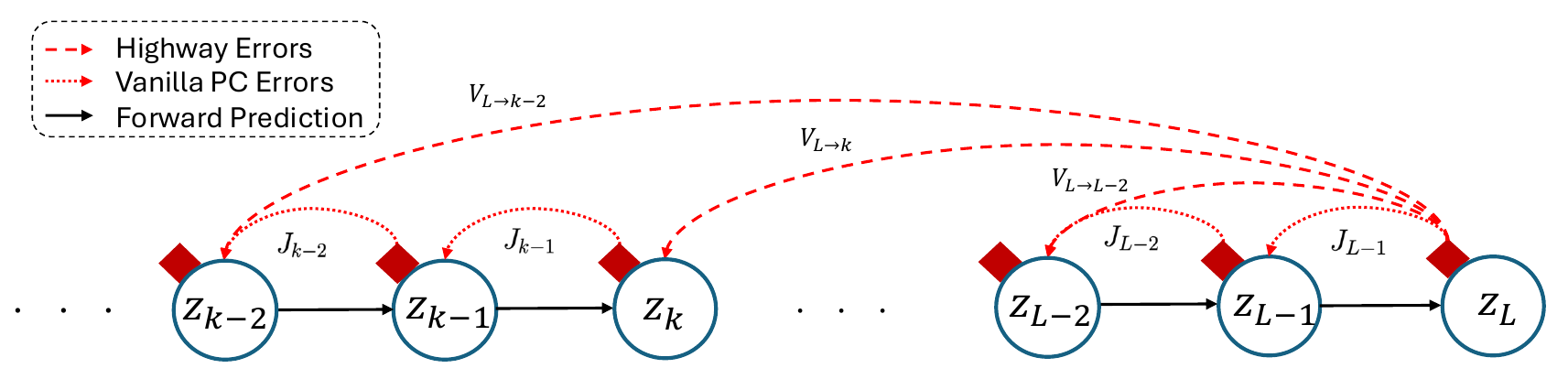}
\caption{\textbf{Our proposed error highways in a deep PCN.} Each (error) highway $V_{\nlayers\to i}$ transmits the output error $\error{\nlayers}$ straight to a selected hidden
layer $i\in S$ within one hop (or E-step), such that the correction applied to that layer does not decay with depth. This stands in contrast to a vanilla PCN, where errors lose a factor of $\lract$ per hop.}
\label{fig:hep_diagram}
\end{figure}

\subsection{The Error-Highway PCN Architecture}
\label{sec:error_highway_structure}

Following the same PCN (backbone) design as introduced in Section~\ref{sec:background}, we have the same states, predictions, and the same clamping of the input and the output (to data and targets, respectively). Inside of our architecture we deliberately do \emph{not} introduce forward skip connections. The only path from input to output prediction runs through every layer, which means that the PCN network cannot route a copy of the signal around untrained layers the way that the $\mu$PC architecture does (Section~\ref{sec:related}).
Any learning that we might observe in the interior layers of the PCN would therefore be attributed to the effect that our feedback/message-passing pathways, which we introduce next.

HEP centers around the structural modification to PC's inference/E-step dynamics through the feedback that reaches the hidden layers (we are motivated further by the positive improvement observed in the marginal log-likelihood of the historical heterarchical generative PC models in \cite{alex_ngc}). We select a set of highway `endpoints' $S\subseteq\{1,\dots,\nlayers-1\}$ and give each endpoint $i\in S$ a direct synaptic projection/connection from the output (specifically, the output's dedicated pair of error units).
The projection, in its simplest form, is a fixed matrix $V_{\nlayers\to i}\in\mathbb{R}^{n_i\times n_{\nlayers}}$, where $n_i$ is the width of layer $i$ and $n_{\nlayers}$ is the dimension of the output error.
We draw/sample its synaptic value entries $V_{\nlayers\to i}$ independently from a standard normal distribution scaled by $\sigma_v$  and never update them, much in the spirit of the fixed random feedback of DFA~\cite{nokland_dfa}.

At every step of inference, each endpoint $i\in S$ is nudged along the direction $V_{\nlayers\to i}\,\error{\nlayers}$, scaled by a single strength coefficient $\alpha>0$, shared by all error highways.
Note that the output error travels along the highway in one hop (i.e., within one E-step). It critically passes through no Jacobian and picks up no factor of $\lract$ along the way; this means that the size of the nudge does not depend on how deep a layer $i$ sits within the PCN structure. This stands in contrast with the $\mathcal{O}(\lract^{\,\nlayers-i})$ signal of Eq.~\ref{eq:transient}.
Moreover, the nudge that affects/perturbs a given hidden layer during the inference is also variable/dynamic during the inference. As the inference closes in on its settling point (equilibrium), the correction shrinks as the $\error{\nlayers}$ gets smaller such that the highway is merely guiding the states rather than overwriting or heavily over-shadowing them.

In an HEP-formulated PCN, the wiring stays local in terms of the biological locality we desire. We remark that a highway endpoint uses only its own state and the broadcasted output error unit values; it does not require information (including any global network information) from the other layers in between (the output and the endpoint) and does not require global backward sweep. This means that our HEP-driven PCN is not locked in terms of any state updates or in terms of any synaptic weight updates \cite{ororbia2023brain}; this satisfies our desired criterion that the M-step Hebbian weight updates of a PCN would proceed as normal/remain unaltered (only the state dynamics require one single dependency - a top-down expectation, a bottom-up error transmission from the nearby layer, and now the output of the recurrent highway that feeds from the output).
We next make HEP's construction precise by writing HEP in terms of an augmented free energy. 

\subsection{The Augmented Free Energy and Inference Dynamics}
\label{sec:energy}

To construct the desired dynamics of each node/layer, we augment the PCN free energy with a bilinear \emph{state-error highway} term between the hidden state $\latent{i}$ and the ``stop-gradient(ed)'' output error:
\begin{equation}
  \label{eq:aug_energy}
  \tilde{F}\!\left(\latents\right)
  =
  \overbrace{E(\latents)}^{F_{\mathrm{pc}} \text{(Energy)}}
  +
  \overbrace{\alpha\sum_{i\in S}\latent{i}^{\!\top} V_{\nlayers\to i}\,\text{sg}(\error{\nlayers})}^{F_{\mathrm{hw}} \text{(Error highway term)}},
\end{equation}
where $E(\latents)$ is the standard PC energy of Eq.~\ref{eq:energy}, and $\text{sg}(\error{\nlayers})$ freezes the output error so the gradient does not flow back into it. This keeps the highway one-directional as the hidden layers receive the output error but cannot alter it in return.
With our augmented energy, we next obtain the gradient flow (E-step) as follows:
\begin{equation}
  \label{eq:z_update}
  \latent{i}^{(t+1)}
  = \latent{i}^{(t)}
  - \lract\bigl(\error{i}^{(t)}-J_i^{(t)\top}\error{i+1}^{(t)}\bigr)
  - \lract\alpha\,V_{\nlayers\to i}\,\text{sg}(\error{\nlayers}^{(t)})\,\mathbb{I}[i\in S].
\end{equation}
Note that, again, for the M-step, the update to the synaptic weights $W_\ell$ remain unchanged (making it the same as in a standard PCN). Since $F_{\mathrm{hw}}$ does not depend on $W_\ell$ (blocked by $\text{sg}$), the
local weight update becomes exactly (Eq.~\ref{eq:W_update_pc}):
\begin{equation}
  \label{eq:W_update}
  \Delta W_\ell
  = -\eta_W\frac{\partial\tilde{F}}{\partial W_\ell}
  = -\eta_W\frac{\partial E}{\partial W_\ell}
  \;\propto\;
  \bigl(\error{\ell+1}\odot\actf'(\cdot)\bigr)\latent{\ell}^{\top}.
\end{equation}

\subsection{Stabilizing the Forward Pass}
\label{sec:normalization}

A practical difficulty in training deep PCNs is ensuring that the neural activities remain under control or well-behaved. Stabilizing the neural activities helps both inference and learning, especially since the prediction errors are the computed differences between the activities and the predictions and the weight updates are themselves scaled by the activities. If the activities grow or shrink as we go deeper, these quantities become unstable and the PCN training breaks down. Therefore, keeping the PCN's forward pass stable is a basic requirement for training deep PCNs, as also argued by~\cite{muPC}.

To implement and enforce a useful normalization in our HEP-PCN, we employ root-mean-squared (RMS) normalization. Before each hidden layer applies its weights to project to the next one, we rescale its input such that its scale no longer depends on the magnitude of the activity. This is formally as follows:
\begin{equation}
  \label{eq:rmsnorm}
  \pred{\ell+1} = \actf\!\big(\synaps_\ell\,\mathrm{RMSNorm}(\latent{\ell})\big),
  \qquad
  \mathrm{RMSNorm}(\latent{\ell}) = \vect{\gamma} \odot \frac{\latent{\ell}}{\mathrm{RMS}(\latent{\ell})},
\end{equation}
where $\mathrm{RMS}(\latent{\ell}) = \norm{\latent{\ell}}/\sqrt{\layerWidth}$ is applied to the $\layerWidth$ hidden units of layer $\ell$ and $\vect{\gamma}$ is a learnable gain ($\phi$ is the transfer function). After this step, only the direction of $\latent{\ell}$ is communicated to the next layer, and its amplitude is divided out.

Note that the initial states of PCN are set by a feedforward propagation pass and this normalization is essential to ensure that the forward pass is stable. Without normalization, the scale compounds from layer to layer, given that each prediction is built from the layer below; this can result in a signal that ``blows up'' or vanishes by the time it reaches the deeper/upstream layers. The RMS norm resets the scale at every layer, ensuring that the predictions $\pred{\ell}$ stay within a controlled scale across the PCN. Bounding the activities also bounds the weight updates. The local PC update of Eq.~\eqref{eq:W_update_pc} is an outer product of the error signal and the pre-synaptic activity. The error is a difference of bounded activities and, if the
pre-synaptic activity is normalized, then the update stays bounded as well. We usefully obtain this property without changing the local form of the update. 

Furthermore, we remark that normalization on its own does not solve the depth problem. It essentially keeps the forward pass and the activities well-behaved; however, the supervisory signal still has to travel back through many layers in order to reach the early ones and, as a result, it weakens along the way (as described in Section~\ref{sec:depthProb}). Ultimately, this is the issue that HEP seeks to resolve; thus, in this work, an HEP-PCN utilizes two important mechanisms: (1) normalization to keep the activities stable, and (2) error highway message-passing pathways to ensure that the error actually reaches the hidden layers.

\section{Experimental Results}
\label{sec:experiments}

We now evaluate HEP in the regime that it was designed for -- deep PCNs with many layers. Our experiments are designed in service of two goals.
First, we measure whether HEP trains deep PCNs where vanilla PC fails, comparing both against backprop (BP) at depths ranging from $4$ to $128$ layers.
Second, we verify that the highway term itself is responsible for this result by sweeping its modulation strength $\alpha$ as well as examining the density of its connections. For all experiments, the architecture is an MLP with no forward skip connections; this again ensures that the only path from input to output runs through every layer (the PCN must use all of its internal layers/states). This deliberate design choice removes the shortcut route around untrained layers that $\mu$PC relies on
(Section~\ref{sec:related}); good accuracy is possible only if interior PCN layers are utilized.


\noindent
\noindent\textbf{Experimental setup.} We train thin MLP-based PCNs (width $\layerWidth{=}128$, ReLU, RMSNorm) at depths $\nlayers\in\{4,8,16,32,64,128\}$ on MNIST and Fashion-MNIST, with cross-entropy on the clamped one-hot output. Two baselines anchor the comparison. \emph{Vanilla PC} is the identical network and training loop with the highway removed ($\alpha{=}0$), so any gap is attributable to the highway alone; \emph{BP} is the same skip-free MLP trained end-to-end. We report test accuracy at the best-validation checkpoint, as mean $\pm$ std over three seeds. Full settings (data splits, per-depth $\alpha$ and $T$, optimizer) are in Appendix~\ref{app:exp_details}.


\begin{table}[t]
\centering
\caption{Test accuracy (\%) versus depth $\nlayers$ on MNIST and Fashion-MNIST}
\label{tab:depth_best}
\small
\setlength{\tabcolsep}{4pt}
\begin{tabular}{llccc}
  \toprule
  & \textbf{$\nlayers$} & \textbf{Vanilla PC} & \textbf{HEP} & \textbf{BP} \\
    \midrule
    \textbf{MNIST}
      & $4$   & $97.8 \pm 0.1$  & $97.8 \pm 0.3 $ & $98 \pm 0.1$\\
      & $8$   & $92.1\pm1.0$  & $96.4\pm0.7$ & $97.8\pm0.1$\\
      & $16$  & $35.5\pm11.2$ & $96.4\pm0.2$ & $97.7\pm0.0$\\
      & $32$  & $13.4\pm2.2$  & $95.6\pm0.3$ & $97.1\pm0.1$\\
      & $64$  & $12.5\pm2.3$  & $95.9\pm0.5$ & $94.9\pm0.2$\\
      & $128$ & $12.8\pm2.5$          & $95.8\pm0.6$ & $95\pm0.1$\\
    \midrule
    \textbf{Fashion-MNIST}
      & $4$   & $88.1\pm 0.1$  & $86.3 \pm 0.5$ & $88.6\pm0.2$\\
      & $8$   & $80.2\pm3.1$  & $83.6\pm0.6$ & $88.6\pm0.2$\\
      & $16$  & $42.2\pm12.4$ & $84.6\pm0.6$ & $88.3\pm0.1$\\
      & $32$  & $17.8\pm3.1$  & $85.1\pm0.1$ & $87.7\pm0.1$\\
      & $64$  & $12.6\pm0.4$  & $83.2\pm0.4$ & $85.4\pm0.0$\\
      & $128$ & $12.4\pm1.4$          & $82.1\pm0.8$ & $85.6 \pm 0.2$\\
    \bottomrule
\end{tabular}
\end{table}
  \begin{figure}[!t]
\centering
\includegraphics[width=0.75\linewidth]{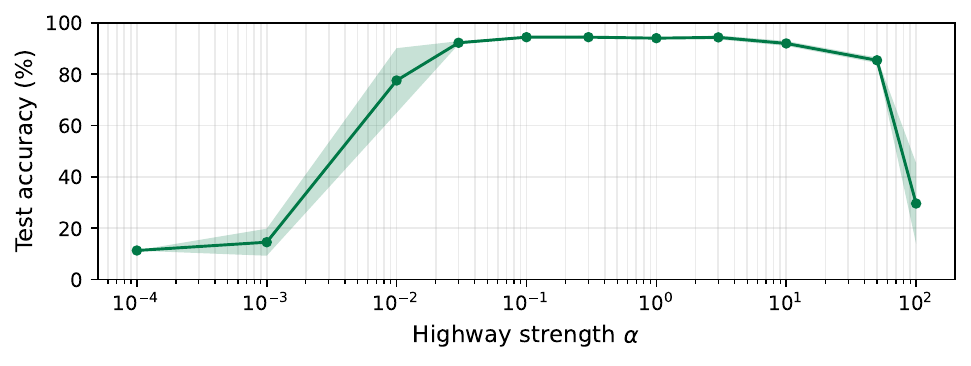}
\caption{Highway strength $\alpha$ vs.\ test accuracy at depth 64 (MNIST).}
\label{fig:alpha-sweep}

\end{figure}
\noindent
\textbf{Accuracy across depth. }
 In Table~\ref{tab:depth_best}, we report the test accuracy from PCN depth $4$ up to depth $128$. At depth $4$, the depth problem has not yet set in. Vanilla PC trains well in this case and HEP offers (naturally) no advantage. However, the picture changes as the network grows/becomes deeper.
Vanilla PC loses ground at depth $8$, partially collapsing at depth $16$, and sits at chance performance from depth $32$ onward. In contrast, HEP stays essentially flat/stable instead -- between $95.6$ and $96.4\%$ on
MNIST and between $82.1$ and $86.3\%$ on Fashion-MNIST -- all the way to $128$ layers. At deeper networks, HEP is comparable to BP on this architecture as
BP itself declines with depth here as well, particularly since the network has no skip connections (and BP suffers from the well-known vanishing gradient problem \cite{pascanu2013difficulty}).
Vanilla PC's poor performance is visible in Figure~\ref{fig:energy-propagation} and, usefully, we observe that the proposed error highways deliver the output error to the interior of the network from the very first inference steps.


\noindent\textbf{Sweeping the highway strength.} If the highway term is what enables learning at depth, turning it off should trace a path back to vanilla PC. At depth 64 on MNIST we sweep $\alpha$ in Eq.~\ref{eq:aug_energy} with every other setting fixed (Figure~\ref{fig:alpha-sweep}). When $\alpha$ is too small the correction is negligible and the network stays near chance; accuracy then climbs to a wide plateau and finally collapses again once $\alpha$ is large enough that the nudges destabilize inference. This transition from chance to plateau, driven by $\alpha$ alone, is our clearest evidence that HEP, and not some other component of the setup, is what enables learning at this depth.

\noindent\textbf{Highway density.} We next ask how sparse the feedback can be. At depth 32 we attach a highway only to every $k$-th hidden layer and vary $k$, keeping the rest of the setup fixed (Table~\ref{tab:everyk}). Thinning is cheap at first, and a coarse covering still trains well. But once the gaps grow too wide, the layers between injection sites stop receiving a usable signal, the attenuation of Section~\ref{sec:depthProb} returns, and individual runs begin to destabilize. A sparse set of highways is therefore enough, which keeps HEP's overhead small.

\begin{table}[h]
  \centering
  \caption{Test accuracy (\%) when only every $k$-th hidden layer receives a highway.}
  \label{tab:everyk}
  \small
  \setlength{\tabcolsep}{3.5pt}
  \begin{tabular}{l cccccc}
    \toprule
    $k$ & $1$ & $2$ & $4$ & $5$ &  $8$ & $10$\\
    \midrule
    Accuracy & $95.6\pm0.3$ & $94.9\pm0.2$ &  $93.1\pm0.1$ & $93.9\pm0.5$ &  $90.3\pm0.7$ & $84.0\pm10.1$\\
    \bottomrule
  \end{tabular}
\end{table}

\section{Conclusion}

\label{sec:conclusion}

In this work, we proposed highway error propagation (HEP), a predictive coding (PC) inference-and-learning scheme, under an augmented free-energy objective, that introduces skip-feedback pathways into the message-passing of PC network (PCN). Our experimental results demonstrate that the proposed HEP successfully trains very deep PCNs, effectively offering a simple, useful solution to the problem of decaying/vanishing error signals when training PC models. Our scheme critically preserves the biological-plausibility of a PCN model, e.g., the neuronal dynamics and synaptic weight updates are local, by further drawing inspiration from the brain's heavy, often heterarchical recurrent connectivity.

\bibliographystyle{acm}
\bibliography{ref}

@article{ororbia2023brain,
  title={Brain-inspired machine intelligence: A survey of neurobiologically-plausible credit assignment},
  author={Ororbia, Alexander G},
  journal={arXiv preprint arXiv:2312.09257},
  year={2023}
}

@article{rumelhart1986learning,
  title={Learning representations by back-propagating errors},
  author={Rumelhart, David E and Hinton, Geoffrey E and Williams, Ronald J},
  journal={nature},
  volume={323},
  number={6088},
  pages={533--536},
  year={1986},
  publisher={Nature Publishing Group UK London}
}

@article{ororbia2022lifelong,
  title={Lifelong neural predictive coding: Learning cumulatively online without forgetting},
  author={Ororbia, Alex and Mali, Ankur and Giles, C Lee and Kifer, Daniel},
  journal={Advances in Neural Information Processing Systems},
  volume={35},
  pages={5867--5881},
  year={2022}
}

@inproceedings{salvatori2024stable,
  title={A stable, fast, and fully automatic learning algorithm for predictive coding networks},
  author={Salvatori, Tommaso and Song, Yuhang and Yordanov, Yordan and Millidge, Beren and Sha, Lei and Emde, Cornelius and Xu, Zhenghua and Bogacz, Rafal and Lukasiewicz, Thomas},
  booktitle={International Conference on Learning Representations},
  volume={2024},
  pages={19607--19631},
  year={2024}
}

@article{bengio2014auto,
  title={How auto-encoders could provide credit assignment in deep networks via target propagation},
  author={Bengio, Yoshua},
  journal={arXiv preprint arXiv:1407.7906},
  year={2014}
}

@inproceedings{resnet,
  title={Deep residual learning for image recognition},
  author={He, Kaiming and Zhang, Xiangyu and Ren, Shaoqing and Sun, Jian},
  booktitle={Proceedings of the IEEE conference on computer vision and pattern recognition},
  pages={770--778},
  year={2016}
}

@article{salvatori2025survey,
  title={A survey on neuro-mimetic deep learning via predictive coding},
  author={Salvatori, Tommaso and Mali, Ankur and Buckley, Christopher L and Lukasiewicz, Thomas and Rao, Rajesh PN and Friston, Karl and Ororbia, Alexander},
  journal={Neural Networks},
  pages={108161},
  year={2025},
  publisher={Elsevier}
}

@article{singer2021recurrent,
  title={Recurrent dynamics in the cerebral cortex: Integration of sensory evidence with stored knowledge},
  author={Singer, Wolf},
  journal={Proceedings of the National Academy of Sciences},
  volume={118},
  number={33},
  pages={e2101043118},
  year={2021},
  publisher={National Academy of Sciences}
}

@inproceedings{muPC,
  title     = {{$\mu$}{PC}: Scaling Predictive Coding to 100+ Layer Networks},
  author    = {Innocenti, Francesco and Achour, El Mehdi and Buckley, Christopher L.},
  booktitle = {The Thirty-ninth Annual Conference on Neural Information Processing Systems},
  year      = {2025},
  url       = {https://openreview.net/forum?id=lSLSzYuyfX}
}

@article{zador2026neuroai,
  title={NeuroAI and Beyond: Bridging Between Advances in Neuroscience and ArtificialIntelligence},
  author={Zador, Anthony and Fellous, Jean-Marc and Sejnowski, Terrence and Adam, Gina and Aimone, James B and Akwaboah, Akwasi and Aloimonos, Yiannis and Alonso, Carmen Amo and Bartolozzi, Chiara and Bennington, Michael J and others},
  journal={arXiv preprint arXiv:2604.18637},
  year={2026}
}

@inproceedings{metaPCN,
title={Stable and Scalable Deep Predictive Coding Networks with Meta-Prediction Errors},
author={Myoung Hoon Ha and Hyunjun Kim and Yoondo Sung and Youngha Jo and Min S. Kang and Sang Wan Lee},
booktitle={The Fourteenth International Conference on Learning Representations},
year={2026},
url={https://openreview.net/forum?id=kE5jJUHl9i}
}

@article{srivastava2015highway,
  title={Highway networks},
  author={Srivastava, Rupesh Kumar and Greff, Klaus and Schmidhuber, J{\"u}rgen},
  journal={arXiv preprint arXiv:1505.00387},
  year={2015}
}

@article{dempster1977maximum,
  title={Maximum likelihood from incomplete data via the EM algorithm},
  author={Dempster, Arthur P and Laird, Nan M and Rubin, Donald B},
  journal={Journal of the royal statistical society: series B (methodological)},
  volume={39},
  number={1},
  pages={1--22},
  year={1977},
  publisher={Wiley Online Library}
}

@inproceedings{zilly2017recurrent,
  title={Recurrent highway networks},
  author={Zilly, Julian Georg and Srivastava, Rupesh Kumar and Koutn{\i}k, Jan and Schmidhuber, J{\"u}rgen},
  booktitle={International conference on machine learning},
  pages={4189--4198},
  year={2017},
  organization={PMLR}
}

@misc{ePC,
      title={ePC: Fast and Deep Predictive Coding for Digital Hardware}, 
      author={Cédric Goemaere and Gaspard Oliviers and Rafal Bogacz and Thomas Demeester},
      year={2026},
      eprint={2505.20137},
      archivePrefix={arXiv},
      primaryClass={cs.LG},
      url={https://arxiv.org/abs/2505.20137}, 
}

@misc{precisionGuided,
      title={Towards the Training of Deeper Predictive Coding Neural Networks}, 
      author={Chang Qi and Matteo Forasassi and Thomas Lukasiewicz and Tommaso Salvatori},
      year={2025},
      eprint={2506.23800},
      archivePrefix={arXiv},
      primaryClass={cs.LG},
      url={https://arxiv.org/abs/2506.23800}, 
}

@article{whittington_bogacz,
    author = {Whittington, James C. R. and Bogacz, Rafal},
    title = {An Approximation of the Error Backpropagation Algorithm in a Predictive
          Coding Network with Local Hebbian Synaptic Plasticity},
    journal = {Neural Computation},
    volume = {29},
    number = {5},
    pages = {1229-1262},
    year = {2017},
    month = {05},
    issn = {0899-7667},
    doi = {10.1162/NECO_a_00949}
}

@article{millidge2022backprop,
    author = {Millidge, Beren and Tschantz, Alexander and Buckley, Christopher L.},
    title = {Predictive Coding Approximates Backprop Along Arbitrary Computation Graphs},
    journal = {Neural Computation},
    volume = {34},
    number = {6},
    pages = {1329-1368},
    year = {2022},
    month = {05},
    issn = {0899-7667},
    doi = {10.1162/neco_a_01497},
    url = {https://doi.org/10.1162/neco_a_01497},
    eprint = {https://direct.mit.edu/neco/article-pdf/34/6/1329/2023477/neco_a_01497.pdf},
}

@article{rao_ballard,
  title   = {Predictive coding in the visual cortex: a functional interpretation of some extra-classical receptive-field effects},
  author  = {Rao, Rajesh P. N. and Ballard, Dana H.},
  journal = {Nature Neuroscience},
  volume  = {2},
  number  = {1},
  pages   = {79--87},
  year    = {1999},
  doi     = {10.1038/4580}
}

@article{friston2005,
    author = {Friston, Karl},
    title = {A theory of cortical responses},
    journal = {Philosophical Transactions of the Royal Society B: Biological Sciences},
    volume = {360},
    number = {1456},
    pages = {815-836},
    year = {2005},
    month = {04},
    issn = {0962-8436},
    doi = {10.1098/rstb.2005.1622},
    url = {https://doi.org/10.1098/rstb.2005.1622},
    eprint = {https://royalsocietypublishing.org/rstb/article-pdf/360/1456/815/89743/rstb.2005.1622.pdf},
}

@article{alex_ngc,
  title = {The neural coding framework for learning generative models},
  volume = {13},
  ISSN = {2041-1723},
  url = {http://dx.doi.org/10.1038/s41467-022-29632-7},
  DOI = {10.1038/s41467-022-29632-7},
  number = {1},
  journal = {Nature Communications},
  publisher = {Springer Science and Business Media LLC},
  author = {Ororbia,  Alexander and Kifer,  Daniel},
  year = {2022},
  month = Apr 
}

@misc{FasterPCN,
      title={Faster Predictive Coding Networks via Better Initialization}, 
      author={Luca Pinchetti and Simon Frieder and Thomas Lukasiewicz and Tommaso Salvatori},
      year={2026},
      eprint={2601.20895},
      archivePrefix={arXiv},
      primaryClass={cs.LG},
      url={https://arxiv.org/abs/2601.20895}, 
}

@article{rec_lra,
  title = {Backpropagation-Free Deep Learning with Recursive Local Representation Alignment},
  volume = {37},
  ISSN = {2159-5399},
  url = {http://dx.doi.org/10.1609/aaai.v37i8.26118},
  DOI = {10.1609/aaai.v37i8.26118},
  number = {8},
  journal = {Proceedings of the AAAI Conference on Artificial Intelligence},
  publisher = {Association for the Advancement of Artificial Intelligence (AAAI)},
  author = {Ororbia,  Alexander G. and Mali,  Ankur and Kifer,  Daniel and Giles,  C. Lee},
  year = {2023},
  pages = {9327–9335}
}

@inproceedings{nokland_dfa,
  author       = {Arild N{\o}kland},
  editor       = {Daniel D. Lee and
                  Masashi Sugiyama and
                  Ulrike von Luxburg and
                  Isabelle Guyon and
                  Roman Garnett},
  title        = {Direct Feedback Alignment Provides Learning in Deep Neural Networks},
  booktitle    = {Advances in Neural Information Processing Systems 29: Annual Conference
                  on Neural Information Processing Systems 2016, December 5-10, 2016,
                  Barcelona, Spain},
  pages        = {1037--1045},
  year         = {2016},
  url          = {https://proceedings.neurips.cc/paper/2016/hash/d490d7b4576290fa60eb31b5fc917ad1-Abstract.html},
  bibsource    = {dblp computer science bibliography, https://dblp.org}
}

@inproceedings{ororbia2019biologically,
  title={Biologically motivated algorithms for propagating local target representations},
  author={Ororbia, Alexander G and Mali, Ankur},
  booktitle={Proceedings of the aaai conference on artificial intelligence},
  volume={33},
  pages={4651--4658},
  year={2019}
}

@article{ororbia2020continual,
  title={Continual learning of recurrent neural networks by locally aligning distributed representations},
  author={Ororbia, Alexander and Mali, Ankur and Giles, C Lee and Kifer, Daniel},
  journal={IEEE transactions on neural networks and learning systems},
  volume={31},
  number={10},
  pages={4267--4278},
  year={2020},
  publisher={IEEE}
}

@article{pessoa2023entangled,
  title={The entangled brain},
  author={Pessoa, Luiz},
  journal={Journal of cognitive neuroscience},
  volume={35},
  number={3},
  pages={349--360},
  year={2023},
  publisher={MIT press One Broadway, 12th Floor, Cambridge, Massachusetts 02142, USA~…}
}

@article{hawkins2025hierarchy,
  title={Hierarchy or heterarchy? a theory of long-range connections for the sensorimotor brain},
  author={Hawkins, Jeff and Leadholm, Niels and Clay, Viviane},
  journal={arXiv preprint arXiv:2507.05888},
  year={2025}
}

@inproceedings{pascanu2013difficulty,
  title={On the difficulty of training recurrent neural networks},
  author={Pascanu, Razvan and Mikolov, Tomas and Bengio, Yoshua},
  booktitle={International conference on machine learning},
  pages={1310--1318},
  year={2013},
  organization={Pmlr}
}

@misc{DKP_PC,
title={Accelerated Predictive Coding Networks via Direct Kolen{\textendash}Pollack Feedback Alignment},
author={Davide Casnici and Martin Lefebvre and Justin Dauwels and Charlotte Frenkel},
year={2026},
url={https://openreview.net/forum?id=MCeZ4k7J6M}
}

@inproceedings{adam,
  author={Diederik P. Kingma and Jimmy Ba},
  title={Adam: A Method for Stochastic Optimization},
  year={2015},
  cdate={1420070400000},
  url={http://arxiv.org/abs/1412.6980},
  booktitle={ICLR (Poster)}}

\newpage
\appendix

\section{The Vanishing Learning Signal: Transient and Steady-State Derivations}
\label{app:depth}

The analyses in this appendix consolidate two complementary results established by recent studies of depth in predictive
coding. The steady-state characterization of Appendix~\ref{app:steady} restates
the classical observation that a PCN reproduces the backpropagation error signal
at its inference equilibrium~\cite{whittington_bogacz,millidge2022backprop}, while
the transient $\lract^{\,\nlayers-i}$ attenuation derived in
Appendix~\ref{app:transient} follows the signal-propagation analysis
of~\cite{ePC}. We restate both derivations here so that the paper is
self-contained and so that the two regimes addressed by HEP are made precise.

We adopt the supervised setting of Section~\ref{sec:background}. The input is
clamped to $\latent{0}\equiv\obs$ and the output to
$\latent{\nlayers}\equiv\target$, leaving the interior states
$\latent{1},\dots,\latent{\nlayers-1}$ free. The supervisory term in the energy of
Eq.~\ref{eq:energy} is the cross-entropy loss
$\ell(\latent{\nlayers},\pred{\nlayers})$, so the clamped output error follows
Eq.~\ref{eq:errors},
\begin{equation}
  \label{eq:app_outerr}
  \error{\nlayers}\;:=\;-\,\nabla_{\pred{\nlayers}}\,\ell\!\left(\latent{\nlayers},\pred{\nlayers}\right),
\end{equation}
the only term that carries target information into the network.

\subsection{Transient regime: the wavefront signal decays as \texorpdfstring{$\lract^{\,\nlayers-i}$}{lambda\^{}(L-i)}}
\label{app:transient}

\emph{Transient analysis} studies how the inference dynamics behave in the very
first steps, before the network reaches equilibrium.
The key question is how quickly the output error signal propagates back through the layers. To show this we will breakdown movement of the latents $\latents$ at beginning of the inference procedure.

Recall the energy gradient for a hidden layer $i=1,\dots,\nlayers-1$,
\begin{equation}
  \label{eq:app_grad}
  \nabla_{\latent{i}}E
  \;=\;
  \error{i}-J_i^{\top}\error{i+1},
  \qquad
  J_i:=\frac{\partial f_i(\latent{i})}{\partial\latent{i}},
\end{equation}
which gives the inference update
\begin{equation}
  \label{eq:app_zupdate}
  \latent{i}^{(t+1)}
  \;=\;
  \latent{i}^{(t)}-\lract\bigl(\error{i}^{(t)}-J_i^{(t)\top}\error{i+1}^{(t)}\bigr),
  \qquad i=1,\dots,\nlayers-1.
\end{equation}
The output layer is clamped, so $\latent{\nlayers}^{(t)}=\target$ for all $t$ and
never updates.

\paragraph{Feedforward initialization.}
Setting $\latent{i}^{(0)}=\pred{i}^{(0)}=f_{i-1}(\latent{i-1}^{(0)})$ for all
$i=1,\dots,\nlayers-1$ makes every hidden prediction error vanish at $t=0$,
\begin{equation}
  \error{i}^{(0)}=0,\qquad i=1,\dots,\nlayers-1.
\end{equation}
The \emph{only} nonzero signal at $t=0$ is therefore $\error{\nlayers}^{(0)}$, the
mismatch at the output layer.

\medskip
\noindent{\color{timecolor}\textbf{Step $t=1$.}}
\begin{itemize}
  \item \textbf{Layer $\nlayers$:}\; clamped at $\latent{\nlayers}=\target$, no update.
  \item \textbf{Layer $\nlayers-1$:}\; since $\error{\nlayers-1}^{(0)}=0$, the update reduces to
  \begin{equation}
    \latent{\nlayers-1}^{(1)}
    = \latent{\nlayers-1}^{(0)}+\lract\,J_{\nlayers-1}^{(0)\top}\error{\nlayers}^{(0)},
    \qquad \Delta\latent{\nlayers-1}=\mathcal{O}(\lract).
  \end{equation}
  \item \textbf{All other layers:}\; remain at their feedforward values; no signal has reached them yet.
\end{itemize}

\medskip
\noindent{\color{timecolor}\textbf{Step $t=2$.}}
\begin{itemize}
  \item \textbf{Layer $\nlayers-1$:}\; because $\latent{\nlayers-2}$ has not moved ($\latent{\nlayers-2}^{(1)}=\latent{\nlayers-2}^{(0)}$), its downstream prediction is unchanged, $\pred{\nlayers-1}^{(1)}=f_{\nlayers-2}(\latent{\nlayers-2}^{(0)})=\latent{\nlayers-1}^{(0)}$. The new error at this layer is therefore
  \begin{equation}
    \error{\nlayers-1}^{(1)}
    = \latent{\nlayers-1}^{(1)}-\latent{\nlayers-1}^{(0)}
    = \lract\,J_{\nlayers-1}^{(0)\top}\error{\nlayers}^{(0)}.
  \end{equation}
  \item \textbf{Layer $\nlayers-2$:}\; since $\error{\nlayers-2}^{(1)}=0$, the update is driven entirely by the error flowing from above,
  \begin{equation}
    \latent{\nlayers-2}^{(2)}
    = \latent{\nlayers-2}^{(1)}+\lract^{2}\,J_{\nlayers-2}^{(1)\top}J_{\nlayers-1}^{(0)\top}\error{\nlayers}^{(0)},
    \qquad \Delta\latent{\nlayers-2}=\mathcal{O}(\lract^{2}).
  \end{equation}
\end{itemize}

\medskip
\noindent{\color{timecolor}\textbf{Step $t=\nlayers-i$.}}
\begin{itemize}
  \item \textbf{General hidden layer $i$:}\; the pattern is clear. Layer $i$ first receives a nonzero signal at time $t=\nlayers-i$, one step later for each layer deeper from the output. At that step layer $i$ has never moved before, so $\error{i}^{(\nlayers-i-1)}=0$ and
  \begin{equation}
    \latent{i}^{(\nlayers-i)}
    = \latent{i}^{(\nlayers-i-1)}+\lract\,J_i^{(\nlayers-i-1)\top}\error{i+1}^{(\nlayers-i-1)}.
  \end{equation}
  By induction the incoming error is $\error{i+1}^{(\nlayers-i-1)}=\mathcal{O}(\lract^{\,\nlayers-i-1})$, so each layer multiplies the signal by one further power of $\lract$,
  \begin{equation}
    \Delta\latent{i}=\mathcal{O}(\lract^{\,\nlayers-i}).
  \end{equation}
\end{itemize}

\begin{resultbox}
Ignoring Jacobian factors, the first backward signal reaching layer $i$ scales as
\begin{equation}
  \label{eq:app_transient}
  \Delta\latent{i}\;\propto\;\lract^{\,\nlayers-i}\, {\error{\nlayers}^{(0)}},
  \qquad \Delta\latent{i}=\mathcal{O}(\lract^{\,\nlayers-i}).
\end{equation}
The initial signal \emph{decays exponentially with depth}: a layer $k$ steps from
the output sees only an $\mathcal{O}(\lract^{k})$ kick on its first update. This is
the attenuation stated in Eq.~\ref{eq:transient}.
\end{resultbox}

\subsection{Steady-state regime: the product-of-Jacobians signal}
\label{app:steady}

\emph{Steady-state analysis} studies the opposite regime: once the inference
dynamics have converged, what does the error profile look like across the layers?
At any stationary point the hidden gradient in Eq.~\ref{eq:app_grad} vanishes,
\begin{equation}
  \error{\ell}-J_\ell^{\top}\error{\ell+1}=0
  \quad\Longrightarrow\quad
  \error{\ell}=J_\ell^{\top}\error{\ell+1},
  \qquad \ell=1,\dots,\nlayers-1.
\end{equation}
Each hidden error is therefore a Jacobian-transpose multiplication of the
next-layer error.  

\paragraph{Closed-form error at depth.}
Unrolling the recursion from layer $i$ up to the output gives
\begin{equation}
  \label{eq:app_steady}
  \error{i}
  = J_i^{\top}J_{i+1}^{\top}\cdots J_{\nlayers-1}^{\top}\,\error{\nlayers}
  = \Bigl(\prod_{k=\nlayers-1}^{\,i}J_k\Bigr)^{\!\top}\error{\nlayers}.
\end{equation}
This is the same product of Jacobians that the chain rule produces when
backpropagation sends the error signal through a feedforward network of depth
$\nlayers$. In other words, at its fixed point the PCN reproduces the backprop
error, and that same product term explains why the error can vanish or explode
as the network gets deeper.

\paragraph{Vanishing and exploding equilibrium error.}
Taking spectral norms in Eq.~\ref{eq:app_steady},
\begin{equation}
  \label{eq:app_specbound}
  \norm{\error{i}}
  \;\le\;
  \Bigl(\prod_{k=i}^{\nlayers-1}\norm{J_k}\Bigr)\norm{\error{\nlayers}}.
\end{equation}
If the Jacobians consistently have spectral norm below one the bound shrinks
geometrically and early layers settle with an exponentially small error. If they stay
above one, the equilibrium errors grow with depth~\cite{metaPCN}.To see this cleanly, restrict to a single direction where every Jacobian acts as a scalar $s$. Then
 \begin{equation}
  \error{i}=s^{\,\nlayers-i}\,\error{\nlayers},
\end{equation}
a pure geometric decay or blow-up in depth for any $\abs{s}\neq1$, with
$\abs{s}=1$ a narrow acceptable regime for the network.

\begin{resultbox}
At equilibrium the error at hidden layer $i$ is the product-of-Jacobians term
\[
  \error{i}=\Bigl(\prod_{k=\nlayers-1}^{\,i}J_k\Bigr)^{\!\top}\error{\nlayers},
\]
the same form that governs vanishing and exploding gradients in feedforward
backprop. Along any direction where the Jacobians act as a scalar $s$,
$\norm{\error{i}}=\abs{s}^{\,\nlayers-i}\norm{\error{\nlayers}}$, so the equilibrium
learning signal decays (or blows up) geometrically in depth. This is the
steady-state counterpart of Eq.~\ref{eq:steady}.
\end{resultbox}

\subsection{Empirical Evidence of Vanishing Learning Signal}

Our goal is to empirically validate the vanishing learning signal phenomena in a deep PCN network. The setup is simple. We
take an \emph{untrained} PCN of 64 layers, clamp a single input--target pair, run the
inference dynamics of Eq.~\ref{eq:app_zupdate} for many relaxation steps, and record the per-layer energy $E_i=\tfrac12\norm{\error{i}}^2$ at every step.
This energy is a direct proxy for the learning signal, since the weight update at layer $i$ scales with $\error{i}$. Wherever $E_i\approx 0$, the layer receives no gradient and
cannot learn.

The relaxation itself is just a minimization of the energy over the free
states, and any optimizer can be used for it. We compare two choices, plain
gradient descent, which is the canonical PC update, and Adam~\cite{adam} as
an alternative.  The gradient-descent case is shown in
Figure~\ref{fig:energy-propagation} and discussed in the main text. In short,
the signal crawls inward one layer per step, loses a factor of $\lract$ at
every hop, and within the inference budget the deep half of the network never
leaves the numerical floor.

\begin{figure}[t]
  \centering
  \includegraphics[width=1\linewidth]{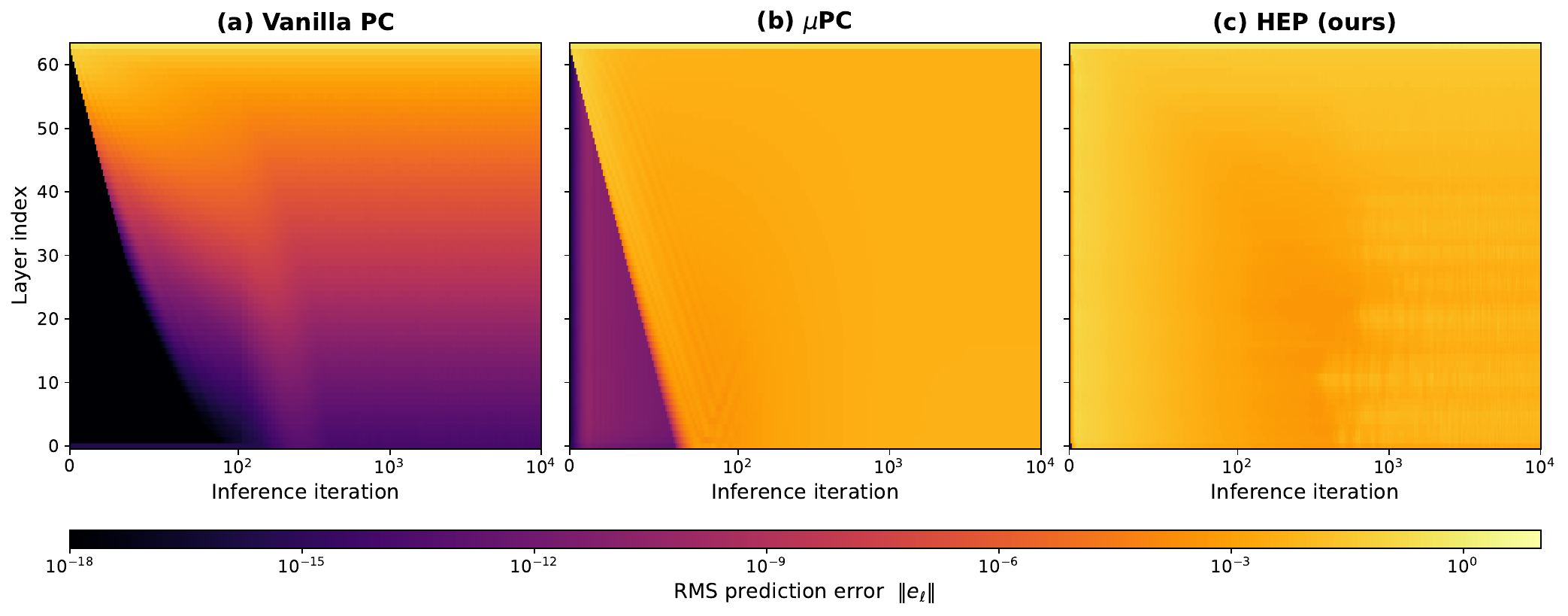}
  \caption{Per-layer RMS prediction error versus layer and inference step for
  the same depth-64 MLP as Figure~\ref{fig:energy-propagation}, with Adam
  ($\epsilon=10^{-2}$) as the inference optimizer. The signal traverses the
  network in far fewer steps than under plain gradient descent and the
  energies settle within the budget.}
  \label{fig:energy-propagation-adam-stable}
\end{figure}

As established in~\cite{muPC}, the inference landscape of a deep PCN is
ill-conditioned, so the descent direction has small components at depth and
plain gradient descent makes negligible progress on them. Adam, on the other
hand, rescales each coordinate by its own running gradient magnitude, which
can magnify tiny directions to a much larger, usable size.
Figure~\ref{fig:energy-propagation-adam-stable} repeats the experiment of
Figure~\ref{fig:energy-propagation} with Adam as the inference optimizer
(with a suitably chosen $\epsilon$, which we return to at the end of this
section). The wavefront now reaches every layer in far fewer steps, and the
per-layer energies settle rather than stalling mid-network.

\begin{figure}[t]
  \centering
  \includegraphics[width=1\linewidth]{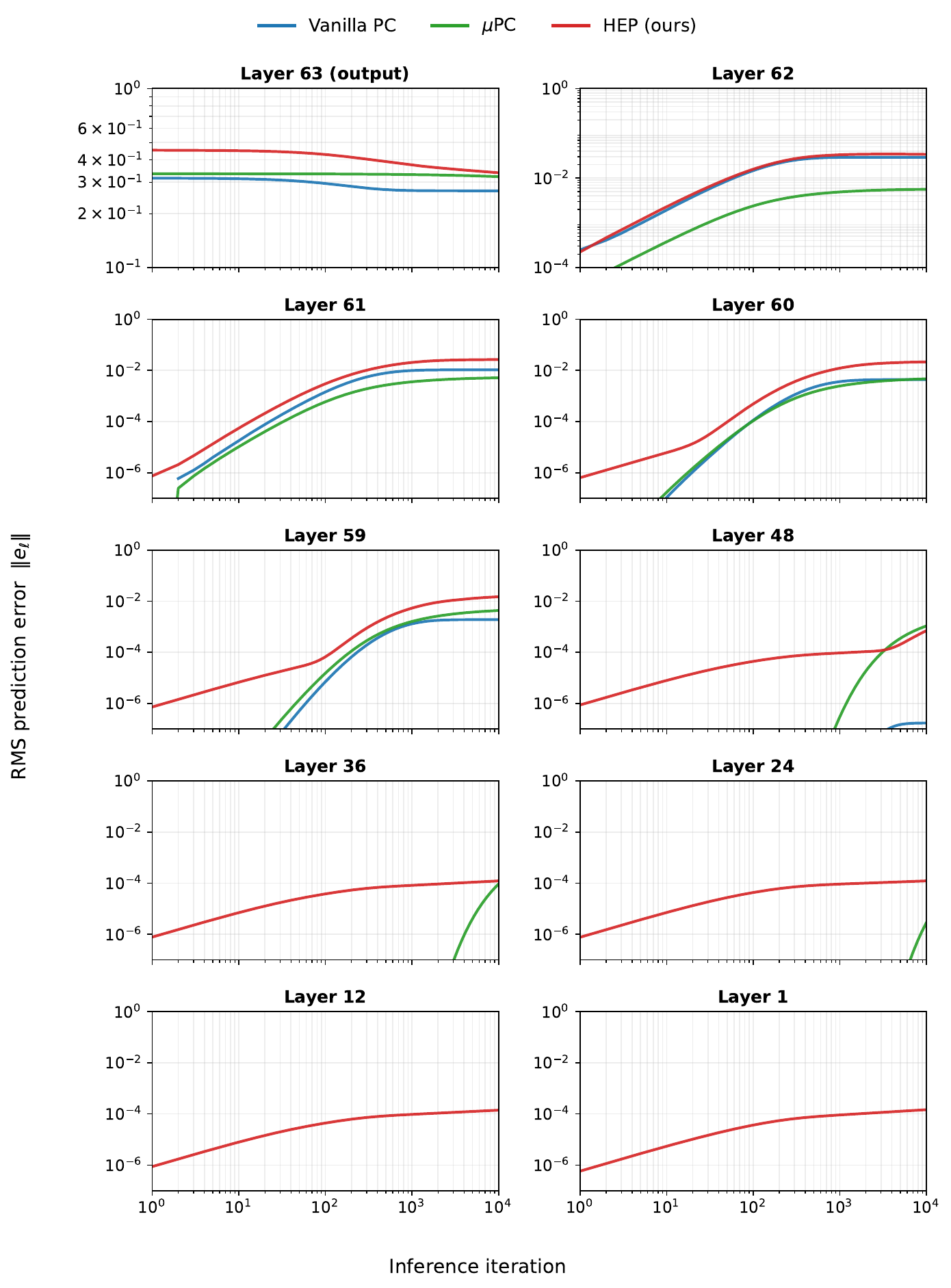}
  \caption{Energy of selected layers over inference steps under plain
  gradient descent. Only the layers next to the clamped output settle; the
  deeper layers keep climbing (or have not activated yet) by the end of the
  run.}
  \label{fig:energy-convergence-gd}
\end{figure}

\begin{figure}[t]
  \centering
  \includegraphics[width=1\linewidth]{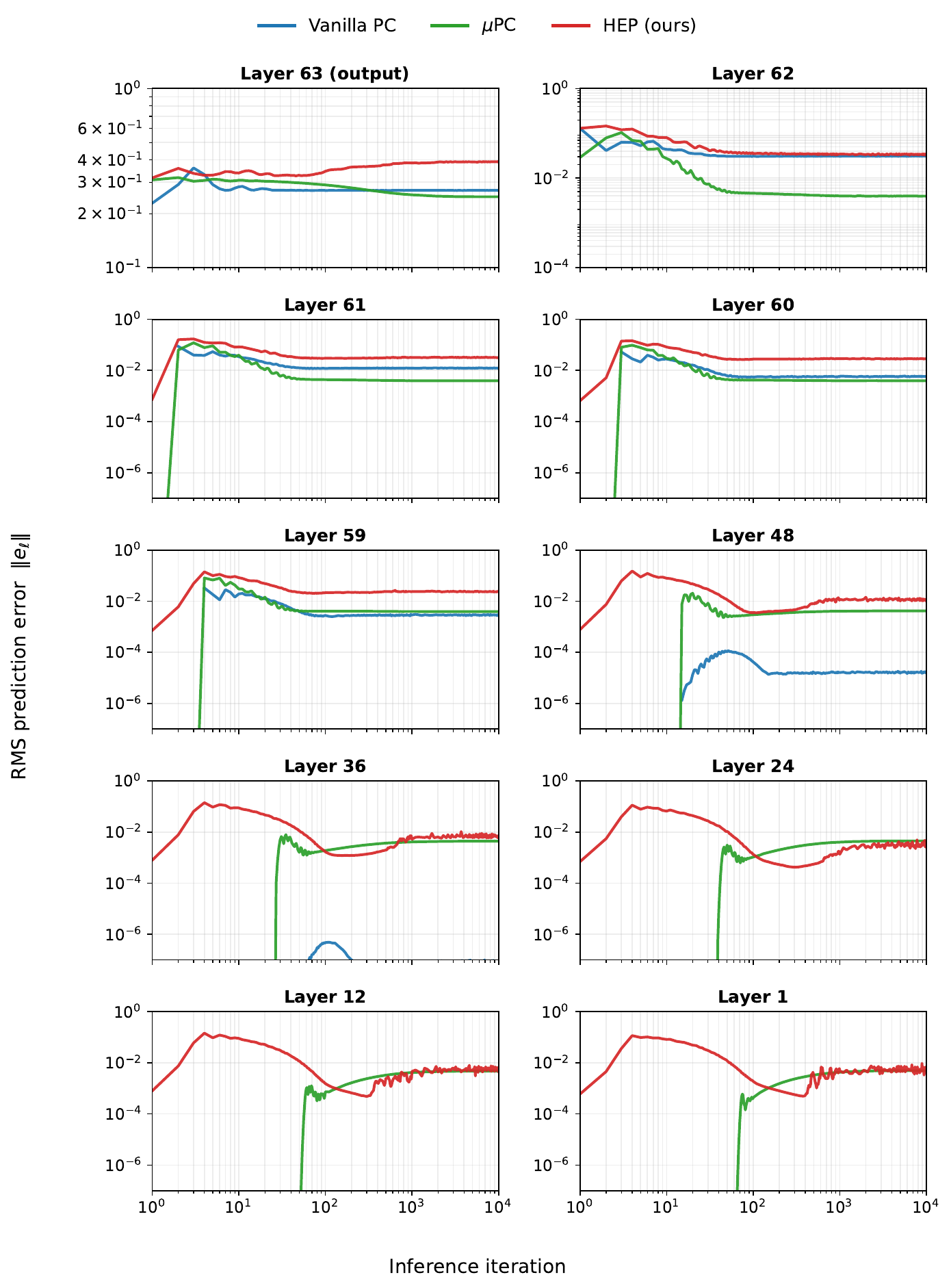}
  \caption{Energy of the same layers under Adam. Every
  tracked layer activates early and its energy levels off at a steady value
  within the budget.}
  \label{fig:energy-convergence-adam}
\end{figure}

To see when each layer receives its signal and whether it settles, we zoom in
on a representative subset of layers. We pick a cluster just below the
clamped output, $\{59,60,61,62,63\}$, and a spread toward the input,
$\{1,12,24,36,48\}$.
Figures~\ref{fig:energy-convergence-gd} and~\ref{fig:energy-convergence-adam}
track the energy of these layers over the inference steps, under gradient
descent and Adam respectively.

Under gradient descent, only the output layer and the few layers directly
below it settle. Every other energy is still rising at the end of the run,
so within $10^4$ steps the inference does not converge
(Figure~\ref{fig:energy-convergence-gd}). Beyond this slow convergence,
whether a layer receives any signal at all depends on its depth. For vanilla
PC, the layers below roughly layer 48 sit at the numerical floor for the
entire inference procedure. Their errors are zero so their
weights are never updated and these layers retain their random initial
mapping. 
$\mu$PC does carry a signal inward as it has forward skip connections, but it reaches the deepest layers only after thousands of steps.
Unlike these methods our method can carry information to early layer in a single hop.
Enabled by feedback connection $V_{\nlayers\to i}$, early layers have
nonzero energy from the very first iteration.

This comparison also clarifies why Adam is important for our method. Even
though the highways nudge every hidden layer from the first iteration, under
gradient descent (Figure~\ref{fig:energy-convergence-gd}) these energies keep rising and do not settle within a finite
budget.
The signal is present everywhere, but the ill-conditioning of the
deep inference landscape makes the descent steps too small for the states to
reach equilibrium. 
Adam removes this obstacle by rescaling each coordinate,
and the energies of our method then settle within the budget (Figure~\ref{fig:energy-convergence-adam}). What Adam cannot do is create a
signal where none arrives, and the deep layers of vanilla PC remain at the
floor even with Adam. The highways guarantee that a genuine signal is
present at every depth, and Adam is what lets the states settle on it within
a practical number of steps.

However, using Adam in the PC setting should be done with care. Not only
should the statistics of Adam be reset at every batch, but the first moment
(the speed) should also be handled carefully. The transient decay (see
App.~\ref{app:transient}) leaves no real signal at the early layers of a
64-layer PCN (e.g.\ $\lract^{62}$ at layer two), so what Adam amplifies is
numerical noise, at a scale set by the constant $\epsilon_{\mathrm{Adam}}$
in its denominator. With the default $\epsilon_{\mathrm{Adam}}=10^{-8}$
this noise is driven to full-size steps and the network appears to activate
everywhere at once. This artifact led us to increase the constant to
$10^{-2}$, which reduces the artifact significantly, although the
genuine yet tiny signal at intermediate depths is still rescaled
(Table~\ref{tab:energy-propagation-optimizer}).

\begin{table}[t]
\centering
\caption{Per-layer error $\norm{\error{\ell}}$ of an untrained depth-64
  vanilla PCN after $10^4$ relaxation steps, for three inference
  optimizers (step size $0.05$). GD shows the expected geometric vanishing;
  Adam with $\epsilon=10^{-8}$ flattens every layer to noise level; raising
  $\epsilon$ to $10^{-2}$ reduces the artifact significantly.}
\label{tab:energy-propagation-optimizer}
\setlength{\tabcolsep}{11pt}
\renewcommand{\arraystretch}{1.1}
\begin{tabular}{r ccc}
\toprule
    Layer $\ell$ & GD & Adam ($\epsilon_{\mathrm{Adam}}{=}10^{-8}$) & Adam ($\epsilon_{\mathrm{Adam}}{=}10^{-2}$) \\
\midrule
$63$          & $2.8\times10^{-1}$  & $2.8\times10^{-1}$ & $2.7\times10^{-1}$ \\
$60$          & $4.0\times10^{-3}$  & $3.3\times10^{-2}$ & $5.7\times10^{-3}$ \\
$50$          & $7.8\times10^{-7}$  & $3.7\times10^{-2}$ & $4.0\times10^{-5}$ \\
$40$          & $2.7\times10^{-10}$ & $3.6\times10^{-2}$ & $4.8\times10^{-7}$ \\
$35$          & $6.1\times10^{-12}$ & $3.3\times10^{-2}$ & $5.5\times10^{-8}$ \\
$30$          & $1.1\times10^{-13}$ & $3.1\times10^{-2}$ & $6.3\times10^{-9}$ \\
$20$          & $2.7\times10^{-17}$ & $3.3\times10^{-2}$ & $5.4\times10^{-11}$ \\
$10$          & ${<}10^{-17}$       & $3.3\times10^{-2}$ & $5.4\times10^{-13}$ \\
$1$           & ${<}10^{-17}$       & $2.2\times10^{-2}$ & $1.8\times10^{-14}$ \\
\bottomrule
\end{tabular}
\end{table}

\section{Experimental Details}
\label{app:exp_details}

This section gives the settings needed to reproduce Section~\ref{sec:experiments}. All runs share the configuration below; only the per-depth knobs in Table~\ref{tab:exp_perdepth} vary. We will release our code upon acceptance.

\paragraph{Data.}
For both MNIST and Fashion-MNIST, we keep the standard $10$k test set and carve a $10$k validation set from the training set, leaving $50$k images for training. Inputs are flattened to $784$ and standardized with per-dataset statistics; targets are one-hot. For each run we report the test accuracy of the checkpoint with the best validation accuracy, aggregated as mean\,$\pm$\,std over seeds $\{0,1,2\}$.

\paragraph{Shared configuration.}
The backbone is an MLP of width $\layerWidth{=}128$ with ReLU, pre-activation RMSNorm (learnable gain, init $1$), and no forward skip connections; weights are initialized as $\Normal(0,\,1/\mathrm{fan\text{-}in})$. The output is hard-clamped to the one-hot target and the supervisory loss is cross-entropy. The free states relax for $T$ steps with Adam-on-states (step $0.005$, default $\beta$s, $\epsilon_{\mathrm{Adam}}{=}10^{-8}$). Importantly, we reset the statistics of the Adam at the beginning of the inference procedure so that the correct statistics be used during the inference.

One weight step then follows with AdamW (weight decay $10^{-4}$). Note that we keep the default $\epsilon_{\mathrm{Adam}}{=}10^{-8}$ here, unlike the larger $10^{-2}$ used for the signal-propagation diagnostic since in HEP training the highways deliver a genuine nonzero signal to every layer from the first inference step, so there is a real signal for Adam to lock onto and the default $\epsilon_{\mathrm{Adam}}{=}10^{-8}$ is safe. The highway matrices $V_{\nlayers\to i}$ are randomly sampled from $\Normal(0,\,\sigma_v^2)$ with $\sigma_v = 10^{-3}$. Batch size is $128$ and training runs for $12$ epochs.

\paragraph{Per-depth settings.}
Only $\alpha$, the inference budget $T$, and the weight learning rate change with depth (Table~\ref{tab:exp_perdepth}); MNIST and Fashion-MNIST use the same values. Depth $4$ is the one exception since we use Euler inference (step $0.5$) at a small budget, due to the fact that smaller networks are not really ill-conditioned.

\begin{table}[t]
  \centering
  \caption{Per-depth HEP hyperparameters (identical for MNIST and
  Fashion-MNIST). The vanilla-PC baseline uses the same rows with
  $\alpha{=}0$.}
  \label{tab:exp_perdepth}
  \small\setlength{\tabcolsep}{8pt}
  \begin{tabular}{rcccl}
    \toprule
    Depth $\nlayers$ & $\alpha$ & $T$ & param.\ lr \\
    \midrule
    $4$   & $0.5$ & $20$  & $1\!\times\!10^{-3}$ \\
    $8$   & $1.8$ & $96$  & $8\!\times\!10^{-5}$ \\
    $16$  & $1.8$ & $96$  & $8\!\times\!10^{-5}$ \\
    $32$  & $0.1$ & $96$  & $6\!\times\!10^{-5}$ \\
    $64$  & $0.1$ & $192$ & $6\!\times\!10^{-5}$ \\
    $128$ & $0.1$ & $384$ & $5\!\times\!10^{-5}$ \\
    \bottomrule
  \end{tabular}
\end{table}

\paragraph{Baselines.}
\emph{Vanilla PC} is the identical network and training loop with $\alpha{=}0$, so any gap is attributable to the highway term alone. \emph{BP} trains the same skip-free MLP end-to-end with backpropagation; its learning rate is swept per depth and selected on validation. At depth $128$ the skip-free net does not converge under the default initialization within the $12$ epoch, so we switch to orthogonal initialization with a warmup-cosine schedule over $24$k steps; this is a change of initialization only.

\paragraph{Ablations.}
Each ablation keeps the per-depth configuration of Table~\ref{tab:exp_perdepth} fixed and varies a single factor. The strength sweep (Fig.~\ref{fig:alpha-sweep}) runs at depth $64$ over $\alpha\in\{10^{-4},10^{-3},0.01,0.03,0.1,0.3,1,3,10,50,100\}$; the density sweep (Table~\ref{tab:everyk}) runs at depth $32$ over $k\in\{1,2, 4,5, 8,10\}$; both use three seeds. The signal-propagation diagnostic (Table~\ref{tab:energy-propagation-optimizer}) uses a single untrained depth-$64$ network with one clamped $(\obs,\target)$ pair, relaxed for $10^{4}$ steps (step $0.05$).

\end{document}